\documentclass{article}

\usepackage[preprint]{neurips_data_2024}

\usepackage[utf8]{inputenc} 
\usepackage[T1]{fontenc}    
\usepackage{hyperref}       
\usepackage{url}            
\usepackage{booktabs}       
\usepackage{amsfonts}       
\usepackage{nicefrac}       
\usepackage{microtype}      
\usepackage{xcolor}         
\usepackage{graphicx}
\usepackage{tabularx}
\usepackage{comment}

\title{Results of the NeurIPS 2023 Neural MMO Competition on Multi-task Reinforcement Learning}

\begin{document}

\author{%
Joseph Su\'arez$^{1*}$ \quad Kyoung Whan Choe$^{2*}$ \quad David Bloomin$^6$ \quad \\ 
\quad \textbf{Jianming Gao}$^{3\dagger}$ \quad \textbf{Yunkun Li}$^{3\dagger}$ \quad \textbf{Yao Feng}$^{3\dagger}$ \quad \textbf{Saidinesh Pola}$^3$ \quad \textbf{Kun Zhang}$^3$ \\ \textbf{Yonghui Zhu}$^3$
 \quad \textbf{Nikhil Pinnaparaju}$^2$ \quad \textbf{Hao Xiang Li}$^2$ \quad \textbf{Nishaanth Kanna}$^2$ \quad \textbf{Daniel Scott}$^2$\\
\textbf{Ryan Sullivan}$^{2,5}$ \quad \textbf{Rose S. Shuman}$^2$ \quad \textbf{Lucas de Alcântara}$^2$ \quad \textbf{Herbie Bradley}$^2$ \\
\textbf{Kirsty You}$^4$ \quad \textbf{Bo Wu}$^4$ \quad \textbf{Yuhao Jiang}$^4$ \quad \textbf{Qimai Li}$^4$ \quad \textbf{Jiaxin Chen}$^4$ \quad\\
\textbf{Louis Castricato}$^2$ \quad \textbf{Xiaolong Zhu}$^4$ \quad \textbf{Phillip Isola}$^1$ \\
$^1$Massachusetts Institute of Technology \quad $^2$CarperAI \quad $^3$Competition Winners \\ \quad $^4$Parametrix.AI \quad $^5$University of Maryland, College Park \quad $^6$Plurality Institute \\
$^{*}$Equal contribution \quad $\dagger$Co-winners\\ 
Correspondence to: \texttt{jsuarez@mit.edu, choe.kyoung@gmail.com}
}
\maketitle

\begin{abstract}

We present the results of the NeurIPS 2023 Neural MMO Competition, which attracted over 200 participants and submissions. Participants trained goal-conditional policies that generalize to tasks, maps, and opponents never seen during training. The top solution achieved a score 4x higher than our baseline within 8 hours of training on a single 4090 GPU. We open-source everything relating to Neural MMO and the competition under the MIT license, including the policy weights and training code for our baseline and for the top submissions.

\end{abstract}

\section{Introduction and Related Work}

Neural MMO 2.0 is a reinforcement learning (RL) platform for massively multiagent research. The environment features 128 agents competing for various resources, items, and equipment in a procedurally generated world with dynamic profession and economy systems. Previous versions have been used in competitions at IJCAI and NeurIPS totaling 1200+ participants and 3500+ submissions.

Related competitions include Procgen 2020 \citep{mohanty2021measuring}, Nethack 2021 \citep{hambro2022insights}, and the MineRL 2020-2023 \citep{ guss2019minerlcomp} and LuxAI 2021-2023 series. Procgen challenges participants to train sample-efficient agents on arcade games. The competition is important for historical purposes as the first large RL competition. Nethack is important for the unprecedented complexity of the environment. The game is an 80s dungeon crawler with procedural generated levels that typically takes new human players months to years to beat for the first time, even if they are using guides and wikis. The MineRL series has proposed multiple challenges over the years, largely focused around sample-efficient learning. These competitions are important for their focus on learning low-level control from pixels, ease of interpretability, and broad appeal.

The LuxAI \citep{Lux_AI_Challenge_S1, lux-ai-season-2} series of multiagent challenges is most closely related to Neural MMO. The first competition had a high emphasis on scripted submissions, but the second competition made RL a primary focus and introduced a new two-player real-time strategy game where each player controls several units. The results are discussed in a YouTube video, but to our knowledge, a full analysis has not been published.

\section{Neural MMO}

\begin{figure}
    \centering
    \includegraphics[width=\linewidth]{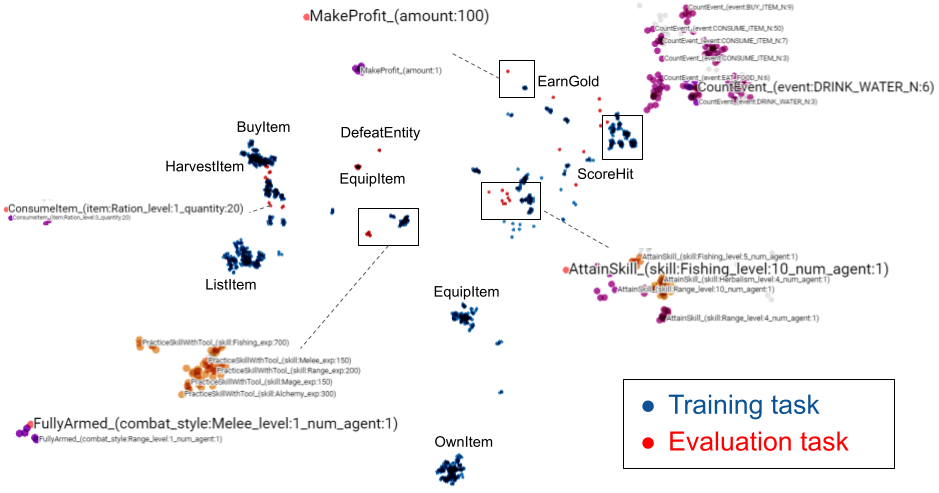}
    \caption{PCA of Neural MMO 2.0 baseline task embeddings. There are 1297 training tasks and 63 evaluation tasks. Similar tasks are grouped into distinguishable clusters. For interactive visualization, see \url{https://kywch.github.io/nmmo-embedding-projector/}} 
    \label{fig:embed}
\end{figure}

Neural MMO has been in active development since 2017 and was first published in 2019 \citep{suarez2019neural}. We refer readers to the recent NeurIPS 2023 D\&B paper \citep{suárez2023neural} for a full description of the environment used in the competition, as well as for comparisons to related environments outside of competitions. For convenience, we briefly summarize the key details of Neural MMO as relates to the most recent competition.

The environment simulates 128 agents on procedural generated maps with several different types of terrain and resources. Agents may move around the map. They must forage for food and water in order to survive and can obtain several other resources. Agents can engage in melee, ranged, or magic combat, which interact with each other in a rock-paper-scissors (or super effective / not very effective) manner. They can also level up foraging and combat related skills to help them perform better. Items like armor, weapons, and ammunition may also be acquired and used to improve combat prowess, while consumables can help agents stay alive. Agents may list items on a global market, and they may in turn purchase items listed by other agents. Suffice to say that Neural MMO simulates a lot of agents and, at least by the standard of environments used in academic research, that the environment supports rich interactions and strategy. At the start of an episode on a new map, each agent is assigned a task to complete. Tasks may involve traveling to a certain location on the map, acquiring certain items or skills, defeating other agents, and so on. Participants are judged by the number of tasks their agents complete over many episodes on different maps.

\section{Task-Conditional Learning}


To our knowledge, this is the first RL competition on \textit{task-conditional} learning. Instead of a single fixed objective, submitted agents are given the task to solve at inference time. This is made possible by Neural MMO 2.0's new task system. The NeurIPS 2023 D\&B paper provides full details, but the key implication for the purpose of this competition is that we can check nearly arbitrary conditions on the agent, environment, and on other agents. The baseline training curriculum has 1297 tasks, including easy goals like eating and drinking. These "scaffolding" tasks are sampled more frequently to enhance agent learning. The evaluation tasks were designed to test if agents successfully engage in all aspects of Neural MMO, spanning survival, combat, exploration, leveling up all skills, using all items, and making money in the market. There are only 63 evaluation tasks, which focus on measuring performance on hard goals. Note that some tasks are too hard to solve without a guiding curriculum. For example, agents perform poorly at the training tasks of equipping medium-level items. Training directly on the evaluation task of equipping high-level items would not improve performance, since agents cannot complete the prerequisites.

Individual tasks are specified in plain code as functions (predicates) called with specific arguments. This means that it is possible to embed tasks into a fixed-length vector by feeding their implementation into a language model. As a result, we have a simple model input that can be used to train task-conditional agents. The training tasks are defined by a set of predicate functions called with specific arguments. When encoded with a LLM, different predicates produce distant embeddings, but the same predicate with different arguments produce similar embeddings. The overlap between training curriculum and evaluation tasks is 22.3 \% when considering only predicates and 2.1 \% when considering both predicate and arguments. As seen in figure \ref{fig:embed}, there is a small but significant gap between most of the training and evaluation task embeddings. Unlike LLM agents, this approach does not require running a large model during inference to control each action. Capable Neural MMO models are only millions to low 10s of millions of parameters.

\section{Rules \& Baseline}

Participants were allowed to modify the model architecture, RL algorithm, and reward function. Training was capped to 8 A100 hours equivalent of training compute and 12 CPU cores. Large hyperparameter sweeps were disallowed, and we informed participants that we would investigate any submission that contained hyperparameters tuned to multiple significant digits. Winners were required to open-source their implementations, and our post-competition investigations (see Discussion) revealed no instances of cheating. We released a 3.9 million parameter baseline model, of which 1 million are contained in the task embedding projection layer. Neural MMO has structured observations and actions that require a structured network architecture to process. A small convolutional subnetwork was used to process observations of the nearby terrain. Linear layers were used to process embeddings of nearby agents as well as items and the market. The action decoder includes a mix of linear layers, for actions such as moving and deciding whether to buy/sell, and pointer networks \citep{vinyals2017pointer}, for selecting targets to attack or an item from the inventory.


The baseline was trained with Clean PuffeRL, which is PufferLib’s variant of CleanRL’s \citep{huang2022cleanrl} PPO \citep{DBLP:journals/corr/SchulmanWDRK17} implementation. The use of PufferLib solves several infrastructure problems, such as batching structured observations and communicating that data across subprocesses. There is also one important modification to the learning algorithm itself that is worth mentioning: because Neural MMO has a variable number of agents, and because most learning libraries (CleanRL included) expect fixed size observations, it is common to zero-pad observations. This is a problem in Neural MMO. It is common for the last quarter of the game to be 90 \% padding because many agents die near the start of the game. This destabilizes learning by lowering and randomizing the effective batch size. Clean PuffeRL alters CleanRL’s data structures to omit padding. This was the key modification that allowed us to improve the usability of our baselines this year. 

\begin{table}
    \caption{Top submissions and baseline performance. PvE refers to preliminary evaluation in an environment without the other participants' submissions. PvP is the final evaluation where all of the top policies are loaded into the same shared environments. The PvE and PvP numbers indicate the completion rate of the evaluation tasks. $\dagger$ indicates the co-winners.}
    \centering
    \begin{tabular}{c ccc}
        \toprule
        \textbf{Team Name} & \textbf{Submission id} & \textbf{PvE (\%)} & \textbf{PvP (\%)} \\
        \midrule
        Takeru$\dagger$    & 246748        & 17.09    & 25.21 \\
        Yao Feng$\dagger$  & 246505        & 16.56    & 24.88 \\
        Saidinesh & 246539        & 10.65    & 12.56 \\
        Mori      & 244278        &  8.84    & 10.61 \\
        Jimyhzhu  & 246707        &  8.14    & 10.07 \\
        Baseline  & n/a           &  5.98    &  6.39 \\
        \bottomrule
    \end{tabular}
    \label{tab:my_label}
\end{table}

\section{Evaluation}

The evaluation consisted of two stages. In the first stage (PvE), all 128 agents were controlled by the submitted policy alone. We ran 32 episodes on the same 4 map seeds, repeating 65 trials per task (32 episode * 128 agents / 63 tasks), which are randomly sampled. As a sanity check, we also evaluated the average agent lifespan. Across 86 submissions, we found a correlation of 0.91 (p<0.001, df=84) between rank according to average lifespan and according to tasks completed. Based on the PvE results, we selected the policies (max 1 per participant) that performed better or comparable to our baseline for the PvP evaluation. Note that the PvE evaluation score can be underestimated if the policy is strong because each agent is evaluated on their own. Strong opponents can impede task performance: the top score in PvE was 17.09, but it was 25.21 in PvP.

In the second round of evaluation, all policies substantially better than the baseline (9 total) were put into the same shared environment. In each episode, 128 agents were split into 9 groups of 14 agents. One of the submitted policies independently controlled all agents within a single group. The remaining two agents were controlled by an untrained baseline policy. Each round of evaluation used 256 held-out maps and 200 episodes, repeating 44 trials per task (200 episodes * 14 agents / 63 tasks), and nine rounds with different seed (i.e., 1800 episodes) were performed to determine winners. We ran multiple evaluations over the course of the competition, and the results were consistent for the same models. We invited the creators of the top submissions to coauthor this manuscript and to document their approaches. Their reports are provided below.

\section{Winning Submissions}

\subsection{"Takeru" by Jianming Gao and Yunkun Li}
\subsubsection{Models and Features}
We observed some instability in the training of the baseline, where the value loss could become excessively large, and the policy entropy would become 0 early. One possible reason for this is that the baseline model did not handle the scaling of continuous features for items properly. We addressed this by scaling all continuous features to [0, 1]. In addition, considering the sparse rewards in difficult tasks and to reduce the difficulty of model training, we made some modification. 
\begin{itemize}
    \item Treat player states, such as coordinates, health, level, as continuous features. 
    \item Remove feature coordinates from Tile states, as the positional information is already modeled using CNN. 
    \item Disable the actions of giving item and giving gold, as these actions are not helpful in the current competition where teamwork is not considered. 
\end{itemize}
This modified baseline model significantly outperforms the original baseline model under the same training configuration. And this is the model of our final solution.

Based on the modified baseline model, we made some attempts to optimize the model structure, but we did not achieve significantly better results. For example, encode player or item entities using Transformer encoder, encode tiles using ResNet blocks. We analyzed that the bottleneck lies in the sparse rewards of difficult tasks, so we didn't invest much time in trying out various model structures.

Due to the constraints of the evaluation setting, we did not invest time in studying the use of LSTM or adding historical information features. However, we believe that historical information is necessary because it allows agent to remember the location of resources, for example.

\subsubsection{Training Configurations}
For the game environment configurations used in training, we conducted some experimental analysis. We have the following different settings compared to the baseline.

\textbf{Number of Maps:} We observed significant differences in task completion rates across different maps. To enhance the model's generalization ability to different maps, we increased the number of maps used for training to 1280.

\textbf{Early Stop Agent Num:} It is used to truncate the game when the number of surviving agents is less than the value of it. The baseline set it to 8 in order to improve sampling throughput. However, we found that it would result in the loss of reward signals for difficult tasks in the later stages of the game, which would hinder learning to complete these difficult tasks. We set it to 0.

\textbf{Resource Resilient:} It is used to reduce the damage suffered by some agents due to starvation or dehydration. It allows agents to survive longer during training, thereby obtaining more reward signals. But we found that it would increase the risk of death of agent under evaluation. We did not use it.
    
\textbf{Number of NPCs:} The baseline increases it from 128 to 256 during training. But we found that the default 128 has enough signals for agents to learn to fight and utilize NPCs. We set it to 128, the same as the evaluation setting.

We also attempted a two-stage training approach. In the first stage, we improved sampling throughput and obtained more reward signals for difficult tasks using these configurations. In the second stage, we continued training under the configurations for evaluation. However, we did not achieve overall better training efficiency and effectiveness. Other game environment configurations for training are identical to the configurations for evaluation. We have not invested time in doing more exploration.

We use PufferLib's Ray parallel sampling on a single machine. The sampling throughput is significantly improved compared to the baseline's Multiprocessing sampling. All hyperparameters of PPO remain the same as the baseline, except for reducing the epochs per training batch from 3 to 1.

\subsubsection{Rewards}
The goal of the game is to complete the task assigned to the agent. By default, the game provides small rewards when there is progress in completing the tasks, in addition to the rewards for task completion. These reward signals are dense enough for easy tasks, such as eating. However, for difficult tasks, such as owning items, the rewards are sparse. We grouped tasks according to their categories and observed that it is challenging to make any progress on difficult tasks.

We studied the 3 custom rewards provided by the baseline. The rewards for restoring health and moving in various directions were unnecessary and the performance was significantly better when these two custom rewards were not used. In contrast, the reward for exploration, which encourages the agent to generate more game events such as causing damage, consuming items, buying items, selling items, etc., had a noticeable positive effect on learning across various tasks. We observed that this custom reward could assist the agent in acquiring general skills, especially in terms of utilizing gold, which would reduce the difficulty of obtaining task progress rewards for difficult tasks. Our final version only utilized the exploration reward and the default rewards provided by the game.

We also attempted to introduce additional custom rewards to enhance the learning of difficult tasks. For example, we attempted to introduce task-specific rewards for attacking or harvesting, as these actions are essential for attaining skill. We also attempted to discourage agent from causing damage to players in order to prolong survival time. However, we did not achieve significantly better results.

\subsection{Yao Feng}
Since the environment is complex and the training time is restricted, our goal is to increase policy expressiveness, decrease exploration difficulty, and help generalization. To achieve these goals, we make changes to the policy, training configurations and the reward function.

\subsubsection{Policy Structure}
Some encoder layers of the baseline did not end in a ReLU. We added ReLU activations to these. Here are other modifications that we made to encoders:

\textbf{Tile Encoder:} We add a ResNet block and use normalized absolute instead of the relative position.

\textbf{Player Encoder:} We remove useless features such as id, attacker\_id and message. Also, we use the same embedding vectors for different attributes to reduce the number of trainable parameters. We also add an MLP over the embedding vectors and layer norm layers before applying activation functions to the outputs.

\textbf{Item Encoder:} We delete the item\_offset property because it is never used.

\textbf{Inventory/Market/Task Encoder:} We add layer norm before applying ReLUs to the outputs.

We add an additional ReLU activation function and two LSTM cells before getting the encoded observation to reduce redundant information and consider the historical information. Because the agent can only see the nearby area, taking historical information into account is important. Besides, we add orthogonal initialization to fully-connected layers. We also restrict the action space to reduce exploration difficulty. Giving items or gold is sub-optimal in most of the cases, so we simply disable these actions. We also disable attacking neutral NPCs, because we observe that this frequently cause the death of our agents.

\subsubsection{Training Configurations}
Here are main changes to the training configurations.

\textbf{PPO Configurations:} We use a learning rate of 1e-4, a clip\_coef of 0.1, and a batch size of 128. We collect about 65,536 steps from 8 environments and save them into one buffer each time before we start training, and we use each sample twice (i.e., ppo\_update\_epochs = 2).

\textbf{Number of Maps:} We increase the number of maps to 1024 to help generalization.

\textbf{Spawn Immunity:} A large value is helpful for initial exploration but not for generalization if the value is not the same in future evaluations. We set it to 20 initially, and the value is selected randomly from $0\sim20$ after the death of any agent.

\textbf{Resilient Population:} This was an experimental training-time setting included with the baseline. We disable it because it is disabled in evaluation.

\subsubsection{Reward Design}
Here are the rewards we used (we skip the task reward because it is given). We do not use the given HP restoration, meander and explore reward because we find them not very useful.

\textbf{HP:} Proportional to changes in HP. Encourages maintaining a healthy status.

\textbf{Experience:} Proportional to the maximum experience across skills. Encourages specialization. 

\textbf{Defense:} Proportional to the average defense (melee, range, mage). Encourages obtaining armor.

\textbf{Attack:} Proportional to the change of the summation of inflicted damage to encourage attacks. 

\textbf{Gold:} Proportional to the change in gold. This discourages buying useless items.

\subsection{Saidinesh Pola}

\subsubsection{Policy Architecture}
In our initial exploration of the NMMO challenge, we identified several shortcomings in the baseline model architecture for centralized training and centralized execution of baseline. Notably, the absence of LSTM layers hindered the model's ability to leverage sequential information from previous observations. Additionally, there was a lack of effective coordination among agents, which we believed could greatly enhance task's performance, even though the task are based individual agent's performance level.

One major limitation we observed was the agents' inability to utilize in-game resources, such as gold, for buying or selling items in the market. Recognizing these deficiencies, we embarked on an extensive investigation, exploring various neural network architectures and techniques to address these issues.

We experimented with different approaches, including cross-attention networks, transformer architectures, and even utilizing ResNet-18, ResNet-34, VIT-tiny. However, we encountered challenges with training stability, prompting us to pivot our focus towards refining the architecture and optimizing model performance with different seeds to avoid zero entropy in earlier steps. For our final submission\footnote{\url{https://github.com/saidineshpola/nmmo-rl}}, we made these modifications to the baseline model architecture:

\textbf{Integration of PyTorch's Multi-Head Attention and Additional Convolutional Layers:} We incorporated multi-head attention mechanisms between agent tiles and introduced an additional convolutional layer to deepen the model's understanding of spatial relationships within the environment.

\textbf{Inclusion of LSTM Layers and Activations:} To capture temporal dependencies and leverage sequential information from past observations, we integrated five LSTM layers into the architecture. This enhancement enabled the model to better understand the evolving dynamics of the game environment and make more informed decisions over time. We adopted the SiLU activation function facilitating smoother gradient flow during training.

\textbf{Multi Head Attention(MHA) Communication Channel:} We utilized cross-attention mechanisms not only for action decoding but also as a communication channel \cite{zhou2023centralized} between agents for better co-ordination. This facilitated the exchange of task-relevant information among agents, ensuring that each agent was aware of the tasks assigned to team. This allowed agents to exchange relevant information about the task at hand, facilitating more coherent and collaborative decision-making processes.


\subsubsection{Reward Tuning and Training}
We encouraged desired behaviors by providing bonuses to the agent's reward: 0.00056 for killing enemies, 0.01 (levels 1-5) or 0.02 (levels above 5) for leveling up combat/fishing skills, 0.03 for giving items or gold to other players, 0.01 for harvesting items, and a variable bonus for meandering (exploring through varied movement patterns) based on the entropy of recent moves. These bonuses incentivize combat, trading, resource gathering, and exploration, shaping the agent's behavior towards engaging objectives to complete different tasks. 

For training, we used the default baseline configuration with 768 agents from 6 environments. To introduce diversity in the training environments, we created 128 new maps for each training run, using different random seeds to generate varied map layouts.

\subsection{"Mori" by Kun Zhang}
Our solution is grounded in the baseline method, with modifications confined exclusively to the reward function and training parameters.

\subsubsection{Model Design Approach}
Due to task variability, we chose to generalize rewards towards improving agent survival, dividing them into two stages—initially emphasizing positive rewards, then shifting to punishment, with manual design followed by genetic algorithm optimization.

Significantly, we initially aimed to optimize TotalScore, but belatedly discovered CompletedTaskCount as the true metric. By the time of this realization, the contest neared its end, so the second-stage design didn't affect the current results. However, it showed promise in boosting TotalScore performance.

\subsubsection{Rewards Design}
Given insignificant progress above the baseline in Curriculum Generation, our research focused on Reinforcement Learning, adopting and empirically validating select reward elements from the previous winner realikun's design. In Stage One, the following reward mechanisms were implemented:

\textbf{Health Recovery Reward:} If health increased due to sufficient food and water intake, a reward of 0.02 was allotted; no penalty was imposed otherwise.

\textbf{Health Point Reward:} A reward proportional to the change in health points (0.005 times the change), providing a 0.05 increment for every 10-point increase and vice versa. This reward was considered effective across all tracks and tasks.

\textbf{Inverted Death Penalty:} Initially intended as a penalty upon agent death, a coding error resulted in a seemingly beneficial effect (+0.02 reward) when the agent died.

\textbf{Task Frequency Reward:} A reward of 0.05 was given for each successfully completed task.

\textbf{Task Progress Reward:} The reward (r) equaled the task completion percentage.

\textbf{Task Completion Reward:} A fixed reward of 3 was awarded for task completion. 

Alongside these, we refined three baseline rewards, setting the event reward to 3 and the exploration reward to 0.5. In Stage Two and Genetic Algorithm Optimization: Due to time constraints, further testing was not conducted. However, employing the Stage Two rewards and genetically optimized reward values led to improved TotalScores and a more stable agent performance, with approximately a 10\% increase observed. The following were found to be ineffective:

\textbf{Food Scarcity Penalty:} Punishment was triggered when food or water levels reached zero.

\textbf{Step-based Rewards:} Phased rewards based on game progression were tested, such as penalizing for food scarcity within the first 500 steps but not thereafter.

\textbf{Poultice Usage Reward:} A reward was for using poultices when the agent had no remaining health.

\subsubsection{Regarding Training Methodology Reflections}
The random nature of the competition highlights the dependence of model training on seed choice. Given the time-consuming and resource-intensive nature of verifying reward settings through multiple model runs, a practical shortcut involves treating seeds as tunable variables and using automatic search algorithms to find advantageous ones. Although this does not inherently enhance the model's abilities, it proves practically useful in boosting scores.

\subsection{"Jimyhzhu" by Yonghui Zhu}
During our exploration of the environment, we identified challenges such as sparse rewards. We believe using historical data will help with the agents decision making; however, due to the constraint on the evaluation system, this is not viable, so we focus on the optimisation of the policy structure.

\subsubsection{Policy Structure}
We identified several weaknesses of the baseline network architecture. This includes the lack of pre-processing of categorical data, scaling issue for continuous data, and lack of global information of other agents in the game. As a consequence, we made the following adjustments:

\textbf{Observation Space Customization:} We expanded the network architecture to incorporate global agents information into Observations(Hidden), which is expected to enhance decision-making by providing a summary of the overall state of other agents in the environment.

\textbf{We introduced one-hot encoding for categorical attributes:} Attributes such as "item type id," and "item equipped status" are transformed into a binary vector representation where the index corresponding to a category's value is set to 1, and all other positions are set to 0. We believe this prevents the model from assuming a natural ordering between categories.

\textbf{Continuous Data Scaling:} The baseline transform continuous data of range [0,100] to a range of [0, 10000] making these values several order of magnitude different from other input. We replaced the baseline's upscaling method, with downscaling, normalizing continuous data to a [0,1] range for balanced feature representation. 

\textbf{Average Pooling reduces sensitivity to order:} We applied average pooling to market embedding and item embedding instead of using a multi layer perceptron as in baseline before combining them into the Observations. This is beneficial because the strategic significance of entities like items or players often does not depend on their order of appearance in the input data.

Actions are decoded using a specific embedding and the the Hidden state, so we decided to experiment with cross attention in the action decoder. The intuition behind this is that by using the Observation as a query, and specific embedding as key and values, the model can selectively extract information from the specific embedding, that are most relevant to the current state as represented by the Observation. However, no major improvement is shown, possibly because the specific embeddings are not rich and diverse enough to allow the cross-attention mechanism to exhibit its full potential. As a consequence cross attention not used in our final design.

\subsubsection{Training Configurations and Reward Design}
For training configurations we increased the number of environments from 6 to 8 and doubled the number of maps from 128 to 256 to improve the model's generalization ability to different maps. We used the default reward function due to limited compute resources and long training time to investigate the effect of reward settings.

\section{Ensuring Fair Evaluations}

We collected source code from the top 5 participants who were eligible for prizes. We inspected their implementations manually and retrained the model from scratch for 8 hours on our hardware. The models were trained between 15M to 22M steps, depending on their speed. This was increased from 10M steps used in the original baseline, as we obtained faster hardware in the interim. Four of the top 5 models matched their original ranks and matched or exceeded original training performance. One model (Mori's) did not quite reproduce, but after investigation, we concluded that this was based on seed instability, not manipulation. In any case, the discrepancy in performance was not so substantial to affect awarding of prizes. We further checked training curves and ensured that everything looked reasonable.

Before determining the final ranks, we gave participants to challenge our evaluation results. A few participants requested and received extended evaluation -- mostly, they wanted to ensure that we had used their best submission and that it performed consistently. There was some contention between the top 2 policies, and after extended evaluation, we declared a tie as co-winners. 

\clearpage
\begin{ack}

Thank you to:
\begin{itemize}
    \item \textbf{CarperAI} for partial sponsorship of compute used for development and for assistance with organization and promotion
    \item \textbf{ParametrixAI} for sponsoring the AICrowd platform fee and for assistance in integrating evaluation with the AICrowd API
    \item \textbf{PufferAI} for sponsoring the prize money and compute used in evaluating and reproducing submissions
\end{itemize}

\textbf{Reduction in scope:} After 6 months of preparation for the competition and well over 1000 hours of work volunteered by contributors, 90\% of our promised compute and devops support was pulled. As a result, we had to functionally eliminate the curriculum track and rewrite our evaluation procedure from scratch while the competition was live. It was a tough situation, but ultimately, I, Joseph, would like to apologize for this failure of management and the additional work it incurred. In order to compensate participants for difficulties these live changes incurred, we manually evaluated checkpoints that failed in our automatic evaluation.

\end{ack}

\bibliographystyle{plainnat}
\bibliography{neurips_data_2023}



\appendix

\section{Supplementary Material}

\begin{itemize}
    \item \textbf{Documentation/Code:} https://neuralmmo.github.io
    \item \textbf{Competition:} https://www.aicrowd.com/challenges/neurips-2023-the-neural-mmo-challenge
    \item \textbf{License:} MIT License
    \item \textbf{Hosting and Maintenance:} The code, documentation, and baselines will continue to be hosted on the Neural MMO Github account. Support is available on the Neural MMO Discord, available from the Documentation page. We will continue to update the platform to resolve major breaking changes.
\end{itemize}

The authors bear all responsibility in the case of violation of rights. Neural MMO is licensed under the MIT License, and the authors confirm that they have the permission to license it as such.

\textbf{Reproducibility:} We provide individual train and run scripts to reproduce the winning policy results in the repository. These may be used as baselines by future works.

\end{document}